\documentclass[twoside]{article}

\usepackage{blindtext} 

\usepackage[sc]{mathpazo} 
\usepackage[T1]{fontenc} 
\usepackage{microtype} 

\usepackage[english]{babel} 

\usepackage[hmarginratio=1:1,top=32mm,columnsep=20pt]{geometry} 
\usepackage[hang, small,labelfont=bf,up,textfont=it,up]{caption} 

\usepackage{enumitem} 
\setlist[itemize]{noitemsep} 

\usepackage{abstract} 

\usepackage{titlesec} 
\renewcommand\thesection{\Roman{section}} 
\renewcommand\thesubsection{\roman{subsection}} 
\titleformat{\section}[block]{\large\scshape\centering}{\thesection.}{1em}{} 
\titleformat{\subsection}[block]{\large}{\thesubsection.}{1em}{} 

\usepackage{fancyhdr} 
\usepackage{titling} 

\usepackage{hyperref} 
\usepackage{cite}
\usepackage[figuresright]{rotating}
\makeatletter
\newif\if@restonecol
\makeatother

\usepackage[linesnumbered,ruled,vlined]{algorithm2e}
\usepackage{algpseudocode}
\usepackage{amsmath}
\usepackage{tabularx}
\usepackage{amssymb}
\usepackage{multicol}
\usepackage{multirow}
\usepackage{url}
\usepackage{authblk}
\usepackage{bbding}

\pretitle{\begin{center}\Huge\bfseries} 
\posttitle{\end{center}} 
\title{An Iterative Path-Breaking Approach with Mutation and Restart Strategies for the MAX-SAT Problem} 
\author[a]{Zhen-Xing Xu \thanks{lxadd515@hust.edu.cn}}
\author[a]{\Envelope Kun He \thanks{brooklet60@gmail.com}}
\author[b]{\Envelope Chu-Min Li \thanks{chu-min.li@u-picardie.fr}}
\affil[a]{Huazhong University of Science and Technology, China}
\affil[b]{MIS, Universit\'{e} de Picardie Jules Verne, France}
\date{\today} 
\renewcommand{\maketitlehookd}{%
\begin{abstract}
\noindent Although Path-Relinking is an effective local search method for many combinatorial optimization problems, its application is not straightforward in solving the MAX-SAT, an optimization variant of the satisfiability problem (SAT) that has many real-world applications and has gained more and more attention in academy and industry. Indeed, it was not used in any recent competitive MAX-SAT algorithms in our knowledge.  In this paper, we propose a new local search algorithm called IPBMR for the MAX-SAT, that remedies the drawbacks of the Path-Relinking method by using a careful combination of three components: a new strategy named Path-Breaking to avoid unpromising regions of the search space when generating trajectories between two elite solutions; a weak and a strong mutation strategies, together with restarts, to diversify the search; and stochastic path generating steps to avoid premature local optimum solutions. We then present experimental results to show that IPBMR outperforms two of the best state-of-the-art MAX-SAT solvers, and an empirical investigation to identify and explain the effect of the three components in IPBMR.
\end{abstract}
}

\begin{document}

\maketitle


\section{Introduction}
\label{introduction}
The Maximum satisfiability problem (MAX-SAT) is an optimization extension of a well-studied typical NP-Complete problem, the satisfiability problem (SAT) \cite{gu1993local,li1997heuristics,gu1999algorithms,audemard2009predicting,cai2012configuration,ansotegui2012community,liang2016exponential,luo2017effective}. They share some aspects but their solvers are very different. The SAT problem is to determine whether there is an assignment of truth values to the propositional variables such that all clauses in a given conjunctive normal form (CNF) formula are satisfied, while the MAX-SAT problem is to find an assignment of truth values such that the number of satisfied clauses is maximized. The importance of different clauses in the MAX-SAT problem can be different. When there are clauses considered to be hard and clauses considered to be soft, the MAX-SAT problem is referred to as the partial MAX-SAT problem, of which the goal is to find an assignment of truth values that satisfies all hard clauses and maximizes the number of satisfied soft clauses. The SAT problem can then be considered as a partial MAX-SAT problem with no soft clauses. When weights are assigned to soft clauses to distinguish their importance, the partial Max-SAT problem is referred to as the weighted partial MAX-SAT problem, of which the goal is to find an assignment of truth values that satisfies all hard clauses while maximizing the total weight of satisfied soft clauses. In recent years, MAX-SAT and its variants have been attracting more and more interests in academy and industry.

As an NP-hard problem, MAX-SAT is very difficult to address and there are two classes of algorithms for solving it: exact algorithms and heuristic algorithms. Exact algorithms (see e.g., \cite{borchers1998two,li2007new,li2010efficient,abrame2014local}) are able to return solutions and prove their optimality. Heuristic algorithms (see e.g., \cite{cha1997local,tompkins2004ubcsat,luo2015ccls,ansotegui2016exploiting,djenouri2017data,ansotegui2017wpm3}) effectively return good-quality solutions that may not be optimal. Searching for the optimal solutions with exact algorithms can be impossible within reasonable time for large instances due to the NP-hardness of the problem. Thus heuristic algorithms, including the local search approaches, are often used for addressing the MAX-SAT problem. The key issue of designing a local search based heuristic algorithm for the MAX-SAT is how to address the local cycling phenomenon in which the search process gets trapped by the local optimum.

As the MAX-SAT problem is very closely related to the SAT problem, one could adapt effective local search strategies for SAT, such as random walk \cite{SKC94}, promising decreasing variable picking \cite{LH05} and configuration checking (CC) \cite{cai2011local,cai2012configuration,abrame2017improving}, to solve the MAX-SAT problem. Unfortunately, making these adaptations effective for MAX-SAT is highly non-trivial because of the difference between SAT and MAX-SAT. In fact, a solution of a SAT instance must satisfy every clause in the instance, meaning that when a clause in the SAT instance is falsified by an assignment, at least one variable in the clause is assigned the wrong truth value. So, picking a variable in the falsified clause and flipping its value to satisfy the clause have a chance to approach a solution of the instance. Therefore, the local search strategies for SAT usually focus on falsified clauses. However, an optimal solution of a MAX-SAT instance can falsify some clauses in the instance, so that flipping the value of any variable in these clauses is a wrong decision. Consequently, guiding the local search using falsified clauses is much more complex for the MAX-SAT than for the SAT.

In this paper, we propose a new strategy named Path-Breaking, instead of falsified clauses, to guide our local search for the MAX-SAT. Path-Breaking is an improved strategy of Path-Relinking method \cite{gloverPR} to adapt to the MAX-SAT. Given two different elite solutions of a combinatorial optimisation problem, Path-Relinking tries to find better solutions by establishing trajectories between the two elite solutions. It has been used to solve many combinatorial optimisation problems \cite{ma2017path,muritiba2018path}, including the MAX-SAT \cite{festa2007grasp}. However, it does not appear to be so effective for the MAX-SAT, because the state-of-the-art local search solvers in the recent MAX-SAT evaluations\footnote{http://maxsat.ia.udl.cat/introduction} do not use Path-Relinking.

In order to make the Path-Relinking method competitive for the MAX-SAT, we identify two drawbacks in the previous Path-Relinking algorithm proposed in \cite{festa2007grasp}: (1) complete trajectories between the two elite solutions are constructed, no matter how the quality of the solutions is in these trajectories, so that many search steps are made in exploring low-quality solutions; (2) the search is not sufficiently diversified, because Path-Relinking is just used to intensify the search around the solutions that have been produced by a GRASP (Greedy Randomized Adaptive Search Procedure) heuristic. Consequently, we propose an effective local search algorithm for the MAX-SAT called IPBMR (Iterated Path-Breaking with Mutation and Restart ) to remedy the above two drawbacks: (1) We establish a condition to break the construction of a trajectory between two elite solutions, allowing the search to focus only on high quality solutions; (2) We randomize the construction of the trajectories between two elite solutions, and if the search falls in a local optimum solution, we perform weak mutations followed by strong mutations that randomly flip a subset of variables of the local optimum solution in order to further diversify the search; (3) We restart \cite{biere2008adaptive} the search to explore new regions of the search space if the mutations do not allow to improve the local mimimum solution.

Our experiments show that IPBMR significantly outperforms the state-of-the-art local search solvers CCLS \cite{CCLS} and Swcca-ms \cite{Swcca-ms}, which do not use Path-Relinking but use falsified clauses to guide the searching process, on most benchmarks in the MAX-SAT evaluation 2016 (MSE2016) \cite{maxsat.ia.udl.cat}. In order to understand the performance of IPBMR, we carry out an empirical investigation to identify and explain the effect of different components of IPBMR.

This paper is organized as follows. Section 2 provides some necessary definitions and notations. Section 3 describes the proposed IPBMR algorithm in detail. Section 4 presents the empirical evaluation of the IPBMR algorithm after describing the experimental environment and the benchmark instances. Section 5 concludes the paper.

\section{Preliminaries}
\label{preliminaries}
A Boolean \emph{variable} $x$ has two possible values: True (denoted as $1$) and False (denoted as $0$). A \emph{literal} $\lnot x$ ($x$) is satisfied if the value $0$ ($1$) is assigned to the variable $x$. It is falsified if the value $1$ ($0$) is assigned to the variable $x$.  A \emph{clause} is a disjunction ($\lor$) of literals and a formula in \emph{conjunctive normal form (CNF)} is a conjunction ($\land$) of clauses.  A clause is satisfied if one of its literals is satisfied and is falsified if all its literals are falsified. A CNF is satisfied if all the clauses are satisfied. A truth assignment assigns a value to each variable in a CNF. For the MAX-SAT problem, any assignment is called a solution of the problem. The MAX-SAT problem is to find an optimal solution that maximizes the number of satisfied clauses, or equivalently minimizes the number of falsified clauses.

Flipping a variable in an assignment is to change its value from 0 to 1 or from 1 to 0. Flipping a variable can make some clauses from falsified to satisfied (Case 1) or from satisfied to falsified (Case 2). The number of clauses in Case 1 is denoted as \emph{make} of the variable and the number of clauses in Case 2 is denoted as \emph{break} of the variable. The \emph{score} of the variable is defined as the value of \emph{make} minus \emph{break}, the net increase in the number of satisfied clauses. The inverse solution of a solution is obtained by flipping all variables.

For example, the CNF formula $\phi=(x_1\lor\lnot x_2)\land(x_2\lor x_3)\land(\lnot x_1\lor x_3)$ contains three clauses.  With the  assignment $\{x_1=0,x_2=1,x_3=1\}$ (denoted as 011), $\phi$ contains two satisfied clauses and one falsified clause. Flipping variable $x_2$, the assignment becomes 001. Then all clauses are satisfied and $\phi$ is satisfied. The \emph{make} of variable $x_2$ is 1. Since no clause is changed from satisfied to falsified after the flip, the break is 0 and the score of $x_2$ is 1.

\section{The IPBMR Algorithm for the MAX-SAT \label{sect-ipbmr}}

In this section we describe the IPBMR algorithm in details. First we modify the Path-Relinking method with a break condition and propose a new algorithm called Path-Breaking (PB). Then, we design an iterative Path-Breaking algorithm with restart strategy (IPBR) based on PB. Finally, we apply a mutation operator borrowed from the Genetic Algorithm \cite{holland1973genetic,holland1992genetic}, to design IPBMR (Iterated Path-Breaking algorithm with Mutation and Restart).

\subsection{The Path-Relinking Method}
Path-Relinking is a strategy used to intensify the search around pair wise elite solutions to find better solutions. For solving the MAX-SAT, a Path-Relinking procedure such as the one used in \cite{festa2007grasp} picks two elite solutions as the starting solution $S_0$ and the target solution $S_t$ respectively, and generates a trajectory $S_0, S_1, S_2, \ldots, S_k=S_t$ from $S_0$ to $S_t$, where $S_i$ $(i=1, 2, \ldots, k)$ is a solution obtained by picking and flipping one variable that has different values in $S_{i-1}$ and the target solution. The best solution along the trajectory is then returned.
Table \ref{Table1} shows a simple example of the Path-Relinking process.

\begin{table}
\caption{A Path-Relinking Process}
\label{Table1}
\centering
\begin{tabularx}{8cm}{cccccc}
\hline
  &$x_1$  &$x_2$ &$x_3$  &$x_4$  &$x_5$ \\
\hline
Starting solution   &$\textbf{1}$  &$0$ &$0$  &$1$  &$0$ \\
                   &$\downarrow$  &  &   &   &  \\
Intemediate solution      &$0$  &$0$ &$0$  &$1$  &$\textbf{0}$ \\
                   &   &  &   &   &$\downarrow$ \\
Intemediate solution      &$0$  &$\textbf{0}$ &$0$  &$1$  &$1$ \\
                   &   &$\downarrow$  &   &   & \\
Target solution    &$0$  &$1$ &$0$  &$1$  &$1$  \\
\hline
\end{tabularx}
\end{table}

In this work, we pick one elite solution as the starting solution and take its inverse solution as the target solution, so that all variables are candidate variables in the beginning. In this way, we explore around one elite solution. So, the search region is bigger in our Path-Relinking process than in the usual Path-Relinking process for the MAX-SAT, because the search region in the usual Path-Relinking process can be considered as the intersection of two regions around the two elite solutions. In fact, the variables having the same value in the two elite solutions are never flipped in a usual Path-Relinking process.

\subsection{The Path-Breaking Strategy \label{Path-Breaking}}

Using the inverse solution as the target solution also makes a longer trajectory and going through this trajectory  needs more calculations, because each variable in the starting solution should be flipped in the complete trajectory. We thus establish a condition to break the search process through the trajectory. At each step of the Path-Relinking process, flipping a variable with positive score improves the solution. Therefore, if there is no positive-score variable, the solution will not be improved.  Fig. 1 contains four figures, each figure showing ten trajectories of the Path-Relinking process for a representative MAX-SAT instance by giving the maximum variable score at each step of the trajectory. These trajectories suggest that, the maximum variable score would become negative after a few steps along the trajectory. We also test a plenty amount of other instances and the trajectories are similar. We thus define a condition to break the Path-Relinking process. Specifically, we record the last positive maximum score (denoted as $lastPos$) and the sum of negative scores after flipping the last variable with positive score (denoted as $sumNeg$). When $\alpha \times lastPos \leq |sumNeg|$, where $\alpha$ is a positive integer parameter, we break the Path-Relinking process, allowing to reduce the amount of calculation by at least half.

 \begin{figure}[ht]
  \centering
  \includegraphics[width=1.0\textwidth]{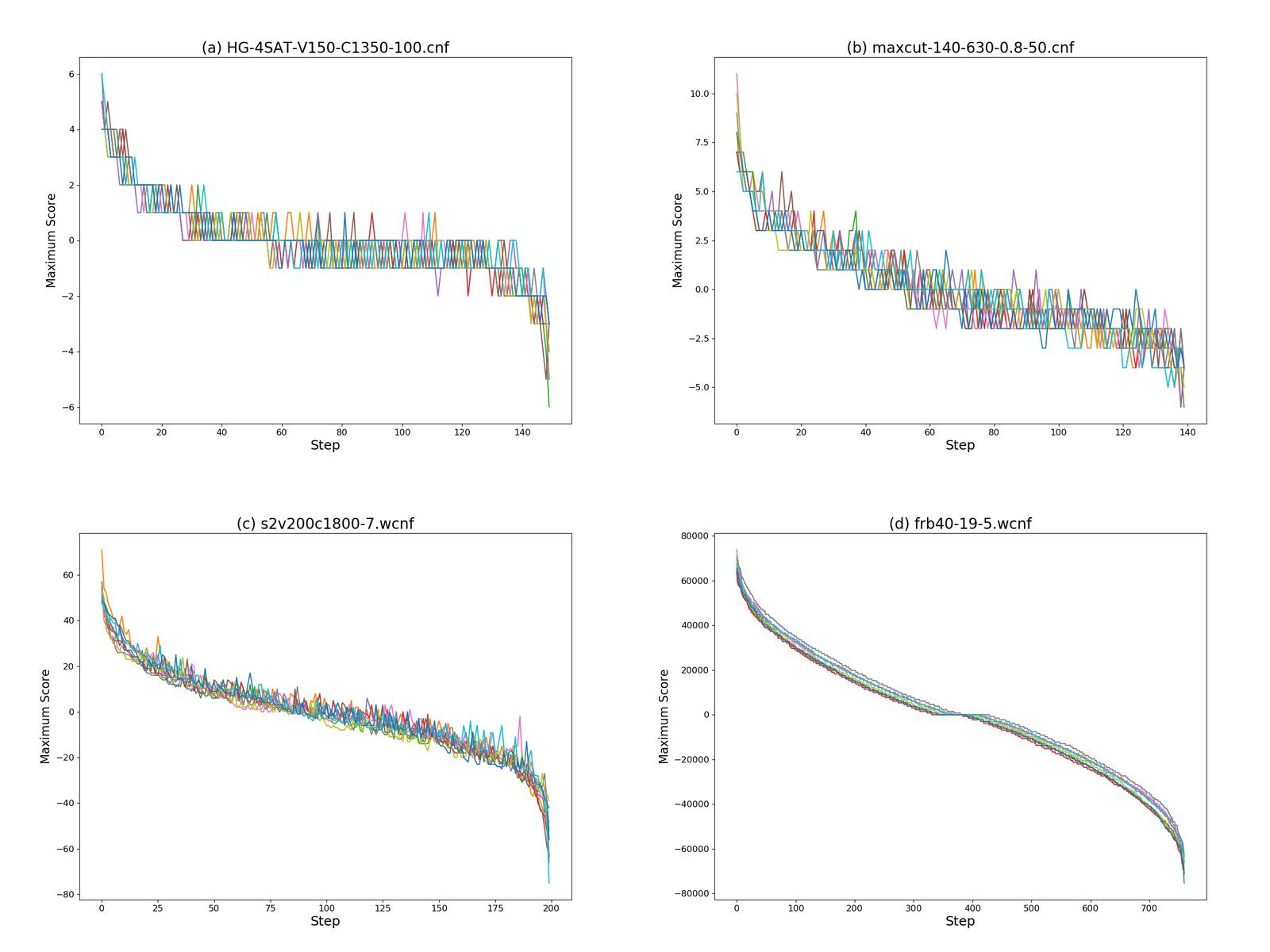}
  \caption{Variation of the maximum variable score in the trajectories for four representative MAX-SAT instances}
  \label{fig.1}
\end{figure}

Besides the break condition, the heuristic to decide which variable to flip at each step is also important. A greedy heuristic is usually used but it often results in premature local optimum solutions. In this paper, we design a heuristic combining greedy method and randomized thought. Algorithm 1 shows the pseudo-code of the Path-Breaking (PB) Algorithm. The details are as follows.

\begin{algorithm}[!tb]
\footnotesize
\caption{\emph{PB($\phi$, $ST$, $\alpha$, $P$): a Path-Breaking algorithm for MAX-SAT}}
\label{PB}
\KwIn{a MAX-SAT instance $\phi$, a starting solution $ST$ of $\phi$, a positive integer parameter $\alpha$, a probability $P$}
\KwOut{a local optimum solution $LOS$ of $\phi$}
$CL \leftarrow$ the set of all variables in $\phi$;  //$CL$ is the set of all candidate variables\\
$LOS$ $\leftarrow$ $ST$; $lastPos \leftarrow 0$;\\
$CS$ $\leftarrow$ $ST$; //$CS$ is the current solution\\
\While {the set of candidate variables $CL$ is not empty} {
     calculate $score(x)$ for each $x \in CL$ \;
     $posVars \leftarrow \{x\ |\ x \in CL\ and\ score(x) > 0\}$;  //$posVars$ is the set of candidate variables with positive score\\
     \If {$posVars$ is empty}{\label{startBreak}
     	$sumNeg \leftarrow \sum_{x \in CL} score(x)$;}
     \Else {
     	$lastPos \leftarrow \max_{x \in posVars} score(x)$;
        $sumNeg \leftarrow 0$;
     }
     \If{$\alpha \times lastPos \leq sumNeg$ }
         {break; //the breaking condition is satisfied \label{endBreak}
         }
     \If {$posVars$ is empty} { \label{startPickVar}
     	$pickedVar \leftarrow argmax_{x \in CL} (score(x))$;
     }
     \Else {
     	with probability $P$,  $pickedVar \leftarrow argmax_{x \in posVars} (score(x))$;\\
	with probability $1-P$,  $pickedVar \leftarrow$ a variable $x$ in $posVars$ selected with probability $\frac{f(x)}{\sum_{y \in posVars} f(y)}$; \label{endPickVar}}
	$CS$ $\leftarrow$ $CS$ with $pickedVar$ flipped;\\
	$CL \leftarrow CL\setminus \{pickedVar\}$;  //remove $pickedVar$ from $CL$ so that it will not be flipped again\\
     \If{ $CS$ is better than $LOS$ } {$LOS \leftarrow CS$\;}
}
\textbf{return} $LOS$;
\end{algorithm}

Given a starting solution \emph{ST}, at each step the PB algorithm flips a picked variable to move along the trajectory towards the inverse solution of \emph{ST}. In order to pick a promising variable, PB calculates the scores for all candidate variables stored in $CL$. We put those variables with positive scores into a list $posVars$ and allocate a probability to be picked to each of them according to their scores. Let $f(x)$ be a function depending on $score(x)$. The probability allocated to  a variable $x$ in $posVars$ is $\frac{f(x)}{\sum_{y \in posVars} f(y)}$. In Algorithm \ref{PB}, we use $f(x)={score^2(x)}$ to give high score variables more opportunity to be flipped.

Concretely, with a constant probability $P$ and if $posVars$ is not empty, PB picks and flips one of the variables in $posVars$ based on their probability. Otherwise (i.e., with probability $1-P$ or if $posVars$ is empty), PB picks and flips a variable with the maximum score among all variables in $CL$. The flipped variable will be removed from the candidate list $CL$. PB continues to pick and flip a candidate variable until the break condition is satisfied or $CL$ becomes empty. Then, the search process is ended and the best solution in the explored part of the trajectory is returned.

\subsection{Iterative PB with Restart Strategy}

We construct an iterative PB algorithm with restart strategy called IPBR as follows. Given a random starting solution $ST$, PB($\phi$, $ST$, $\alpha$, $P$) returns an improved solution, then we take this solution as the starting solution $ST$ for the next PB process. But the iteration will be stopped when the current $ST$ cannot be improved any more, which happens when there is no positive scored variable at the beginning of a PB process. Then, the iteration is restarted by generating  a new random starting solution $ST$ and by iteratively improving $ST$ using the PB process. Finally, The best solution found in these iterations is returned. The pseudo-code of the iterative algorithm IPBR is presented in Algorithm \ref{IPBR}.

Note that due to the random generation of the starting solution, each restart allows IPBR to explore a different search region.

\begin{algorithm}[!tb]
\footnotesize
\caption{\emph{IPBR($\phi$, $\alpha$, $P$, $maxNbRestarts$), an iterative Path-Breaking algorithm with restart}}
\label{IPBR}
\KwIn{A MAX-SAT instance $\phi$, a positive integer parameter $\alpha$, a probability $P$ and a maximum number of restarts $maxNbRestarts$}
\KwOut{A solution of $\phi$}
Generate a random solution $RS$ of $\phi$;\\
\While{$k \leftarrow 1$ to $maxNbRestarts$}
{
     Generate a random starting solution $ST$ \;
     $CS \leftarrow ST$; //$CS$ is the current solution that will be iteratively improved using Path-Breaking\\
     \While{ $true$ }
     {
       $LOS \leftarrow PB(\phi, CS, \alpha, P)$ \;
       \If{ $LOS$ is better than $CS$  } {$CS \leftarrow LOS$\;}
       \Else {$break$ \;}
     }
    \If{ $CS$ is better than $RS$  } {$RS \leftarrow CS$\;}
}
\textbf{return} $RS$;
\end{algorithm}

\subsection{The IPBMR Algorithm}
To further improve the IPBR algorithm, we adapt and apply the mutation operator from the genetic algorithm to IPBR. In fact, when the search process reaches a local optimum, IPBR just restarts the iteration to search in other regions. However, in a  neighbourhood of the local optimal solution larger than the one that can be reached in a Path-Breaking process, there may exist a better solution. Let $CS$ be the current local optimal solution. We randomly flip some variables in $CS$ to obtain a new solution that is not too far from $CS$ and maintains a part of the good quality of $CS$. Then we improve the new solution using the Path-Breaking algorithm.

The IPBMR (Iterative Path-Breaking with Mutation and Restart) algorithm is the IPBR algorithm with the mutating and improving strategy described above, as depicted in  Algorithm \ref{IPBMR}. In the IPBMR algorithm, the mutation acts on the local optimum solution returned by each iterative Path-Breaking process. We exploit two types of mutation. The strong mutation is to flip a large percentage of variables and the weak mutation is to flip a small percentage of variables. The percentage is set with a prerequisite to maintain the major structure of the local optimal solution. Thus we set 70\% for the strong mutation and 20\% for the weak mutation. Both mutations are operated $maxNbMutations$ times, where $maxNbMutations$ is a parameter, to increase the chance of finding a better solution. The search continues from the mutated local optimum solution.

\begin{algorithm}[!th]
\footnotesize
\caption{\emph{IPBMR($\phi$, $\alpha$, $P$, $maxNbRestarts$, $maxNbMutations$),  an iterative Path-Breaking algorithm with mutation and restart}}
\label{IPBMR}
\KwIn{A MAX-SAT instance $\phi$, a positive integer parameter $\alpha$, a probability $P$, a maximum number of restarts $maxNbRestarts$ and a maximum number of mutations $maxNbMutations$}
\KwOut{A solution of $\phi$}
Generate a random solution $RS$ of $\phi$;\\
\For{ $i \leftarrow 1$ to $maxNbRestarts$ }
{
     Generate a random starting solution $ST$ \;
     $CS \leftarrow ST$; //$CS$ is the current solution that will be iteratively improved using Path-Breaking\\
	$CBS \leftarrow ST$; //$CBS$ is the best solution so far in the current restart\\
	$nbWeakMutations \leftarrow 0$;  $nbStrongMutations \leftarrow 0$; \\
     \While{$true$}
     {
       $LOS \leftarrow PB(\phi, CS, \alpha, P)$ \;
       \If{ $LOS$ is better than $CS$  } {
			$CS \leftarrow LOS$\;
			\If {$LOS$ is better than $CBS$} {
                   $CBS \leftarrow LOS$;
		     	$nbWeakMutations \leftarrow 0$;  $nbStrongMutations \leftarrow 0$; \\
              }
	  }
       \ElseIf  {$nbWeakMutations < maxNbMutations$} {
			$CS \leftarrow CBS$; $nbWeakMutations++$;\\
			 \For{each variable $x$ in $CS$} {
	    		flip $x$ in $CS$ with probability 0.2;}

       }
	  \ElseIf {$nbStrongMutations < maxNbMutations$}{
          $CS \leftarrow CBS$; $nbStrongMutations++$;\\
		 \For{each variable $x$ in $CS$} {
	    		flip $x$ in $CS$ with probability 0.7;}
				
       }
       \Else  {$break$ \;}
    }
    \If{ $CBS$ is better than $RS$  } {$RS \leftarrow CBS$\;}
}
\textbf{return} $RS$;
\end{algorithm}

\section{Experiments and Discussions}

The IPBMR algorithm is presented in Section \ref{sect-ipbmr} for the unweighted MAX-SAT for convenience. In reality, we also implemented it for the weighted MAX-SAT and the weighted partial MAX-SAT (recall that the unweighted MAX-SAT is a special case of the weighted MAX-SAT in which the weight of each clause is 1). For the weighted MAX-SAT, the {\em make (break)} of a variable is the total weight of falsified (satisfied) clauses that will become satisfied (falsified) if the variable is flipped. For the weighted partial MAX-SAT, if there are falsified hard clauses, the {\em make (break)} of a variable is the number of falsified (satisfied) hard clauses that will become satisfied (falsified) if the variable is flipped; otherwise, the {\em make (break)} of a variable is the total weight of falsified (satisfied) soft clauses that will become satisfied (falsified) if the variable is flipped. In any case, the score of a variable is equal to $make-beak$.

In the implementation of IPBMR, the parameters are empirically fixed as follows. The parameter $\alpha$ in the breaking condition is set to 3, and the probability $P$ to pick the variable with the greatest score in the PB funciton is 0.2 for the weighted MAX-SAT and to 0.99 for the weighted partial MAX-SAT instances. The maximum number of mutations $maxNbMutations$ applied to a local minimum solution is set to 3 for industrial instances and to 7 for the other instances.  The parameter $maxNbRestarts$ is set to $+\infty$, as the cutoff time is set to 300 seconds.

We conducted two experiments to evaluate IPBMR. In the first experiment, we compared IPBMR with two state-of-the-art local search algorithms for the MAX-SAT that do not use the Path-Relinking strategy, but focus on falsified clauses to pick the next variable to flip. In the second experiments, we compared IPBMR with three variants of IPBMR to analyze the performance of IPBMR.

The benchmarks used in the experiments come from the MSE2016 and were divided into two groups: unweighted MAX-SAT and weighted partial MAX-SAT. Each group consisted of three types of instances: Random, Crafted and Industrial. IPBMR was implemented in C programming language and all experiments were performed on an Intel(R) Xeon(R) CPU E5-2640 v4 @ 2.40GHz under Linux operating system.

In the experiments, each solver solved each instance 10 times independently with different random seeds, and each run was limited to 300 seconds and 4 GB of memory. Let $t_i$ ($1\le i \le 10$) be the time in seconds in which a solver  found the best solution in the $i$th run for an instance. The solving time of the solver for the instance is defined as $(t_1+t_2+\cdots+t_{10})/10$. The instances are divided into subset according to their size and/or their properties as in the MSE2016, in which the number of instances is denoted as ``\#inst.". We compared the solvers for each subset of instances, in terms of the total solving time in seconds for the instances of the subset, denoted as ``time"; and the number of instances for which a solver reported the best solution (i.e., the solution satisfying the greatest number of clauses) among the compared solvers, denoted as ``\#win.". A solver is better than another for a subset of instances if it reported better solution for more instances (i.e. with a greater ``\#win") in the subset, tie being broken using the shorter total solving time.

\subsection{Comparing IPBMR with two state-of-the-art solvers}

We compared IPBMR with the following two state-of-the-art solvers which are among the best in MSE2016:

\begin{description}
\item[{\em CCLS}]\cite{CCLS} : It is an efficient local search algorithm for the MAX-SAT which applies the configuration checking with the {\em make} heuristic to pick and flip variables. The configuration checking (CC) strategy is a kind of tabu method that forbids a variable $x$ to be flipped again after it is flipped, until a variable occurring in a clause containing $x$ is flipped.  The score of a variable is equal to the {\em make} of the variable in CCLS, where {\em make} is defined as in IPBMR. A step of CCLS can be described as follows.
With a fixed probability, CCLS randomly chooses a falsified clause $c$ and flips a randomly chosen variable in $c$. Otherwise, let $C$ be the set of the variables in falsified clauses that are not forbidden by the CC strategy. If $C$ is not empty, CCLS flips the variable in $C$ with the greatest {\em make}; otherwise, CCLS randomly chooses a falsified clause $c$ and flips a randomly chosen variable in $c$.
\item [{\em Swcca-ms}]\cite{Swcca-ms}: It is an adaptation of the local search algorithm for SAT called Swcca \cite{swcca} to MAX-SAT. In Swcca-ms, the weight of each clause is initiated to 1. Then, every time Swcca-ms encounters a local minimum solution, the weight of each falsified clause is incremented by one. The score of a variable is the total weight of the falsified clauses that will become satisfied minus the total weight of the satisfied clauses that will become falsified if the variable is flipped. A step of Swcca-ms can roughly described as follows.  Let $C$ be the set of the variables with positive score that are not forbidden by the CC strategy (see the above description of CCLS). If $C$ is not empty, flip the variable in $C$ with the greatest score. Otherwise, if there are variables with very great score that are forbidden by the CC strategy, flip such a variable with the greatest score. Otherwise, randomly choose a falsified clause $c$ and flip the least recently flipped variable in $c$.
\end{description}

CCLS and Swcca-ms are available on the homepage of the MSE2016. The results in the MSE2016 show that the two solvers are competitive. Note that the variables with positive {\em make} or positive score occur necessarily in falsified clauses. The search strategy of CCLS and Swcca clearly focuses on falsified clauses. The comparison of IPBMR with CCLS and Swcca is in reality a comparison of a search strategy guided by Path-Relinking with a strategy focusing on falsified clauses.

Table \ref{Table2} shows the results of CCLS, Swcca-ms and IPBMR on the 454 unweighted random instances and the 402 crafted instances of the MSE2016.From Table \ref{Table2}, we can see that, IPBMR is better than CCLS and Swcca-ms for almost all subsets of instances, the total solving time being divided by 10 or more for some subsets.

\renewcommand{\arraystretch}{1.5}
\begin{table}[!tb]

  \centering
  \fontsize{8}{8}\selectfont
  \caption{Comparisons on Unweighted MAX-SAT Benchmarks}
  \label{Table2}
  \setlength{\tabcolsep}{4mm}{
    \begin{tabular}{|c|c|c|c|c|}
    \hline
    \multirow{2}{*}{Ins.Type}&\multirow{2}{*}{Subset Name (\#inst.)}&
    CCLS&Swcca-ms&IPBMR\cr
    &&avg. time   (\#win.)& time   (\#win.)&time   (\#win.)\cr
    \hline
    \multirow{10}{*}{Random}
    &120v(45)   &0.09(45)	&1.28(45)	&{$<$\bf0.01(45)} \cr\cline{2-5}
    &140v(45)	&0.56(45)	&0.72(45)	&{$<$\bf0.01(45)} \cr\cline{2-5}
    &160v(45)	&0.38(45)	&0.75(45)	&{$<$\bf0.01(45)} \cr\cline{2-5}
    &180v(44)	&1.25(44)	&0.56(44)	&{\bf0.02(44)} \cr\cline{2-5}
    &200v(49)	&0.12(49)	&0.49(49)	&{\bf0.10(49)} \cr\cline{2-5}
    &70v(45)	&0.81(45)	&0.98(45)	&{$<$\bf0.01(45)} \cr\cline{2-5}
    &90v(49)	&0.59(49)	&0.91(49)	&{$<$\bf0.01(49)} \cr\cline{2-5}
    &110v(50)	&2.59(50)	&0.95(50)	&{\bf0.06(50)} \cr\cline{2-5}
    &3sat(50)	&165.86(50)	&{\bf8.09(50)}	&115.60(50) \cr\cline{2-5}
    &4sat(32)	&166.57(32)	&{\bf5.47(32)}	&30.02(32) \cr\cline{2-5}
    \hline
    \multirow{11}{*}{Crafted}
    &maxcut-140-630-0.7(50)	&0.16(50)	&0.98(50)	&{\bf0.13(50)} \cr\cline{2-5}
    &maxcut-140-630-0.8(50)	&{$<$\bf0.01(50)}	&0.77(50)	&{$<$\bf0.01(50)} \cr\cline{2-5}
    &v140(45)	&0.54(45)	&296.54(45)	&{\bf0.14(45)} \cr\cline{2-5}
    &v160(45)	&4.70(45)	&7.64(45)	&{\bf0.33(45)} \cr\cline{2-5}
    &v180(45)	&1.11(45)	&15.51(45)	&{\bf0.52(45)} \cr\cline{2-5}
    &v200(45)	&4.65(45)	&92.40(45)	&{\bf0.82(45)} \cr\cline{2-5}
    &v220(45)	&2.05(45)	&22.79(45)	&{\bf0.99(45)} \cr\cline{2-5}
    &dimacs-mod(62)	&0.84(62)	&0.54(62)	&{$<$\bf0.01(62)} \cr\cline{2-5}
    &spinglass(5)	&0.07(5)	&0.07(5)	&{$<$\bf0.01(5)} \cr\cline{2-5}
    &scpclr(4)	&{\bf0.14(4)}	&312.95(4)	&225.31(1) \cr\cline{2-5}
    &scpcyc(6)	&{\bf305.37(6)}	&196.79(1)	&391.57(1) \cr\cline{2-5}
    \hline
    \end{tabular}}
\end{table}

Table \ref{Table3} shows the results of CCLS, Swcca-ms and IPBMR on the weighted partial random and crafted MAX-SAT instances of the MSE2016. The Swcca-ms reported false solution for most of these instances because the reported total weight of the falsified clauses is not equal to the sum of the weights of the clauses falsified by the reported solution. IPBMR is significantly better than CCLS on these instances.

\renewcommand{\arraystretch}{1.5}
\begin{table}[!tb]

  \centering
  \fontsize{8}{8}\selectfont
  \caption{Comparisons on Weighted Partial Random and Crafted Benchmarks}
  \label{Table3}
  \setlength{\tabcolsep}{4mm}{
    \begin{tabular}{|c|c|c|c|c|}
    \hline
    \multirow{2}{*}{Ins.Type}&\multirow{2}{*}{Subset Name (\#inst.)}&
    CCLS&Swcca-ms&IPBMR\cr
    &&avg. time (\#win.)&time (\#win.)&time (\#win.)\cr
    \hline
    \multirow{12}{*}{Random}
	&120v(50)	&{$<$\bf0.01(50)}	&False	&{$<$\bf0.01(50)}\cr\cline{2-5}
	&140v(50)	&0.15(50)	&False	&{\bf0.10(50)}\cr\cline{2-5}
	&160v(45)	&0.33(45)	&False	&{\bf0.32(45)}\cr\cline{2-5}
	&180v(44)	&0.44(44)	&False	&{\bf0.34(44)}\cr\cline{2-5}
	&200v(49)	&1.22(49)	&False	&{\bf0.89(49)}\cr\cline{2-5}
	&70v(45)	&0.38(45)	&False	&{\bf0.15(45)}\cr\cline{2-5}
	&90v(49)	&1.00(49)	&False	&{\bf0.30(49)}\cr\cline{2-5}
	&110v(50)	&2.03(50)	&False	&{\bf1.21(50)}\cr\cline{2-5}
	&wpmax2sat/hi(30)	&311.65(29)	&False	&{\bf7.85(29)}\cr\cline{2-5}
	&wpmax2sat/lo(30)	&250.84(29)	&False	&{\bf2.09(30)}\cr\cline{2-5}
	&wpmax2sat/me(30)	&85.34(29)	&False	&{\bf4.93(30)}\cr\cline{2-5}
	&wpmax3sat/hi(30)	&1.50(30)	&False	&{\bf0.03(30)}\cr\cline{2-5}
    \hline
    \multirow{14}{*}{Crafted}
    &auc-paths(20)	&{\bf1.3(20)}	&False	&80.13(20)\cr\cline{2-5}
	&auc-scheduling(50)	&{$<$\bf0.01(20)}	&False	&1.33(20)\cr\cline{2-5}
	&CSG(10)	&868.27(2)	&False	&{\bf710.74(6)}\cr\cline{2-5}
	&dimacs\_mod(43)	&0.03(43)	&False	&{$<$\bf0.01(43)}\cr\cline{2-5}
	&frb(34)	&{\bf539.19(34)}	&False	&1927.05(17)\cr\cline{2-5}
	&miplib(12)	&{\bf165.16(3)}	&False	&44.35(1)\cr\cline{2-5}
	&ramsey(15)	&{\bf590.03(15)}	&False	&899.81(8)\cr\cline{2-5}
	&random\_net(32)	&{\bf5667.22(30)}	&False	&4646.64(2)\cr\cline{2-5}
	&scp4x(10)	&1367.67(2)	&False	&{\bf1274.52(9)}\cr\cline{2-5}
	&scp5x(10)	&1642.85(2)	&False	&{\bf1191.79(8)}\cr\cline{2-5}
	&scp6x(5)	&{\bf293.63(3)}	&False	&647.2(3)\cr\cline{2-5}
	&scpn(20)	&2490.19(7)	&False	&{\bf1665.47(18)}\cr\cline{2-5}
	&spinglass(5)	&84.04(4)	&False	&{\bf112.48(5)}\cr\cline{2-5}
	&warehouses(18)	&188.85(6)	&False	&{\bf2865.5(12)}\cr\cline{2-5}
    \hline
    \end{tabular}}
\end{table}

The performance of local search solvers generally is not good enough for weighted partial industrial MAX-SAT instances, as is showed in recent MAX-SAT evaluations. Table \ref{Table4} shows the results of CCLS, Swcca-ms and IPBMR on the weighted partial industrial MAX-SAT instances of the MSE2016, for which at least one of the three compared solvers reported a solution satisfying all hard clauses for at least one instance in the subset. Swcca-ms also reported false solutions on most instances, indicating that it is not suitable for weighted partial instances.The results show that IPBMR is substantially better than CCLS for these instances, indicating that the Path-Relinking strategy can substantially improve the performance of local search for industrial instances.

\renewcommand{\arraystretch}{1.5}
\begin{table}[!tb]

  \centering
  \fontsize{8}{8}\selectfont
  \caption{Comparisons on Weighted Partial Industrial Benchmarks}
  \label{Table4}
  \setlength{\tabcolsep}{4mm}{
    \begin{tabular}{|c|c|c|c|c|}
    \hline
    \multirow{2}{*}{Ins.Type}&\multirow{2}{*}{Subset Name (\#inst.)}&
    CCLS&Swcca-ms&IPBMR\cr
    \multirow{14}{*}{Industrial}
    &&avg. time (\#win.)&time (\#win.)&time (\#win.)\cr
    \hline
	&BTBNSL(60)	&N/A	&False	&{\bf544.93(4)}\cr\cline{2-5}
	&correlation-clustering(129)	&1039.26(1)	&False	&{\bf3736.2(20)}\cr\cline{2-5}
	&dir(21)	&1400.58(9)	&False	&{\bf2216.64(16)}\cr\cline{2-5}
	&packup-wpms(99)	&153.08(1)	&False	&{\bf2155.96(13)}\cr\cline{2-5}
	&haplotyping-predigrees(100)	&{\bf1993.62(23)}	&False	&N/A\cr\cline{2-5}
	&log(21)	&{\bf1522.14(21)}	&False	&2285.61(5)\cr\cline{2-5}
	&preference\_planning(29)	&{\bf22.35(6)}	&False	&N/A\cr\cline{2-5}
	&relational-inference(9)	&N/A	&False	&{\bf247.23(1)}\cr\cline{2-5}
    &upgradeability-problem(100)	&N/A	&False	&{\bf12906.39(98)}\cr\cline{2-5}
    \hline
    \end{tabular}}
\end{table}

\subsection{Performance analysis of IPBMR}

In order to analyze what makes the performance of IPBMR, we compared IPBMR with three variants of IPBMR described as follows.

\begin{description}
\item[\em IPBMR-noRandom: ]It is IPBMR, unless the PB function (Algorithm \ref{PB}) systematically flips the variable with the greatest score in each Path-Relinking step. In other words, the lines \ref{startPickVar}-\ref{endPickVar} in Algorithm \ref{PB} are replaced with $pickedVar \leftarrow argmax_{x \in CL} (score(x))$.
\item[\em IPBMR-noBreak: ]It is IPBMR, unless the PB function (Algorithm \ref{PB}) systematically builds a complete trajectory without breaking it. In other words, the lines \ref{startBreak}-\ref{endBreak} in  Algorithm \ref{PB} are removed.
\item[\em IPBR: ]It is Algorithm \ref{IPBR}. The difference between IPBR and IPBMR is that IPBR does not use any mutation when it encounters a local optimum solution, while IPBMR does, as Algorithm \ref{IPBR} and Algorithm \ref{IPBMR} show.
\end{description}

Table \ref{Table5} shows the results of IPBMR-noRandom, IPBMR-noBreak, IPBR and IPBMR on unweighted MAX-SAT benchmark of MSE2016. From Table \ref{Table5}, we can see that, IPBMR and IPBMR-noRandom are substantially better than IPBR and IPBMR-noBreak for all subsets, meaning that the mutations of local optimum solutions and the Path-Breaking strategy in IPBMR and IPBMR-noRandom are essential for their performance. IPBMR is slightly better than IPBMR-noRandom, meaning that some randomness in the choice of the variable to flip is useful, especially for the less random instances such as {\em 3sat} and {\em scplr}, because a greedy choice usually leads to a local minimum solution more easily.

\renewcommand{\arraystretch}{1.5}
\begin{table}[!tb]

  \centering
  \fontsize{8}{8}\selectfont
  \caption{Comparisons in different variants of IPBMR on Unweighted MAX-SAT Benchmarks}
  \label{Table5}
  \setlength{\tabcolsep}{2.5mm}{
    \begin{tabular}{|c|c|c|c|c|c|}
    \hline
    \multirow{2}{*}{Ins.Type}&\multirow{2}{*}{Subset Name (\#inst.)}&
    IPBMR-noRandom &IPBMR-noBreak &IPBR&IPBMR\cr
    &&avg. time   (\#win.)&avg. time   (\#win.)&avg. time   (\#win.)&avg. time   (\#win.)\cr
    \hline
    \multirow{10}{*}{Random}
    &120v(45)   &0.01(45)	&0.43(45)	&0.01(45) &{$<$\bf0.01(45)}\cr\cline{2-6}
    &140v(45)	&0.01(45)	&1.02(45)	&0.11(45) &{$<$\bf0.01(45)}\cr\cline{2-6}
    &160v(45)	&0.03(45)	&1.39(45)	&0.07(45) &{$<$\bf0.01(45)}\cr\cline{2-6}
    &180v(44)	&0.04(44)	&3.09(44)   &0.18(44) &{\bf0.02(44)}\cr\cline{2-6}
    &200v(49)	& \bf 0.07(49)	&4.94(49)   &0.37(49) &{0.10(49)}\cr\cline{2-6}
    &70v(45)	&0.01(45)	&0.53(45)	&0.06(45) &{$<$\bf0.01(45)}\cr\cline{2-6}
    &90v(49)	&0.06(49)	&1.67(49)	&0.15(49) &{$<$\bf0.01(49)}\cr\cline{2-6}
    &110v(50)	&{\bf0.05(50)}	&4.89(50)	&0.64(50) &0.06(50)\cr\cline{2-6}
    &3sat(50)	&233.39(50)	&4316.69(24) &2703.54(44) &{\bf115.60(50)}\cr\cline{2-6}
    &4sat(32)	&50.12(32)	&1052.92(30) &516.08(32) &{\bf30.02(32)}\cr\cline{2-6}
    \hline
    \multirow{11}{*}{Crafted}
    &maxcut-140-630-0.7(50)	&0.17(50)	&5.63(50)  &1.32(50)  &{\bf0.13(50)}\cr\cline{2-6}
    &maxcut-140-630-0.8(50)	&0.24(50)  &6.51(50)  &1.2(50)  &{$<$\bf0.01(50)} \cr\cline{2-6}
    &v140(45)	&0.38(45)	&6.63(45)	&1.56(45) &{\bf0.14(45)}\cr\cline{2-6}
    &v160(45)	&0.77(45)	&12.65(45)	&4.25(45) &{\bf0.33(45)}\cr\cline{2-6}
    &v180(45)	&\bf 0.52(45)	&14.59(45)	&6.74(45) &{\bf0.52(45)}\cr\cline{2-6}
    &v200(45)	&2.00(45)	&43.93(45)	&17.1(45) &{\bf0.82(45)}\cr\cline{2-6}
    &v220(45)	&1.84(45)	&79.37(45)	&22.02(45) &{\bf0.99(45)}\cr\cline{2-6}
    &dimacs-mod(62)	&$<$\bf0.01(62)	&0.02(62)	&{$<$\bf0.01(62)} &{$<$\bf0.01(62)}\cr\cline{2-6}
    &spinglass(5)	&0.43(5)	&5.18(5)	&1.76(5) &{$<$\bf0.01(5)}\cr\cline{2-6}
    &scpclr(4)	&759.21(3)	&550.13(1)	&353.41(0) &{\bf225.31(3)}\cr\cline{2-6}
    &scpcyc(6)	&392.24(4)	&426.16(1)	&176.46(1) &{\bf391.57(4)}\cr\cline{2-6}
    \hline
    \end{tabular}}
\end{table}

\begin{figure}[!th]
  \centering
  \includegraphics[width=0.9\textwidth]{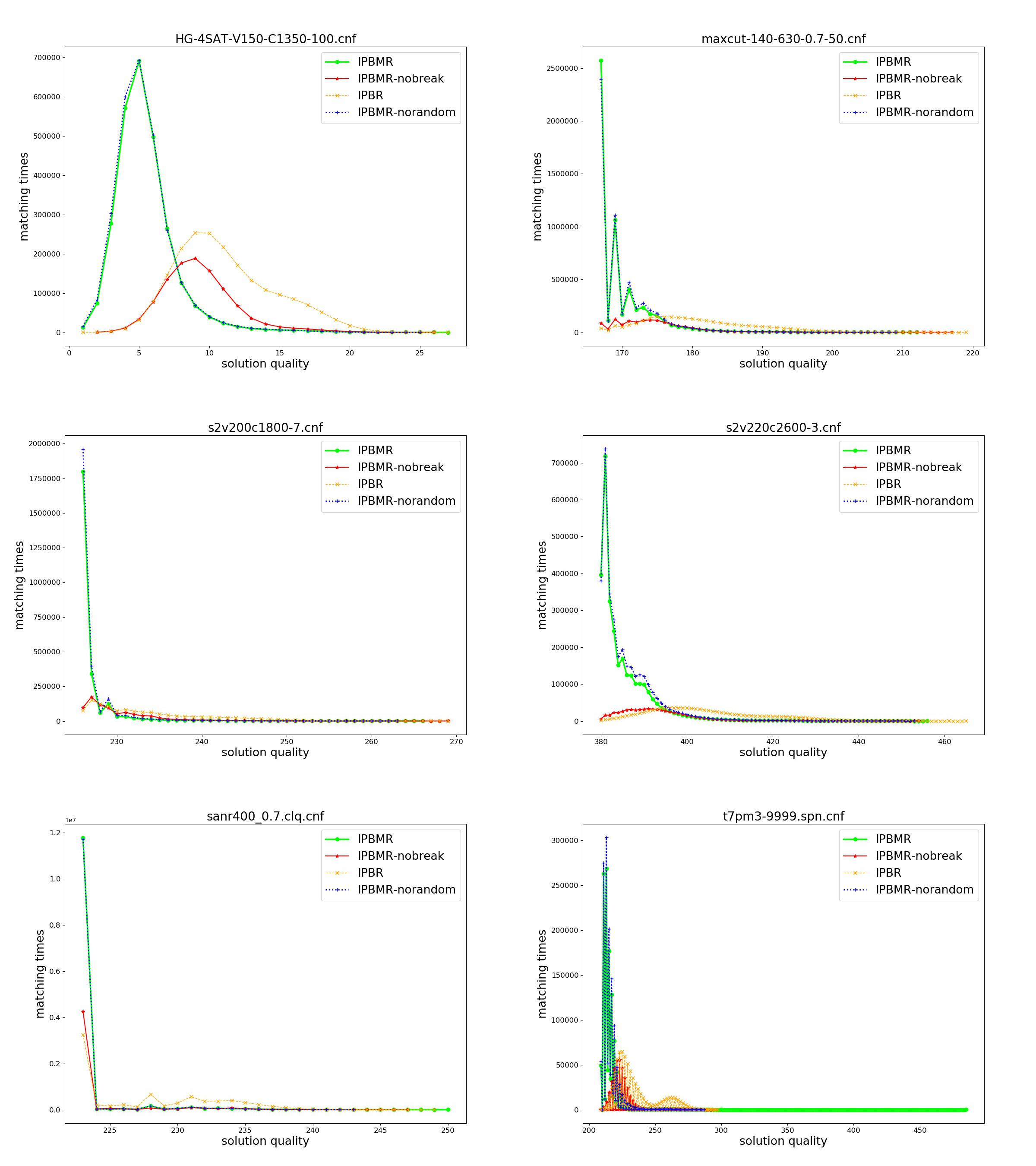}
  \caption{Distribution of unsatisfied clauses for 6 representative MAX-SAT instances}
  \label{fig.2}
\end{figure}

We now show how the mutations of local optimum solutions and the Path-Breaking strategy improve the performance of IPBMR and IPBMR-noRandom. Recall that IPBMR, IPBR, IPBMR-noBreak and IPBMR-noRandom all iteratively call the PB function (Algorithm \ref{PB}), each call of the PB function returning a solution which falsifies some clauses. We ran each of the four solvers to solve six representative instances in the unweighted benchmarks of the MSE2016, using a cutoff time of 300 seconds for each instance; and collect the set of solutions returned by the PB function, which we partitioned according to the number of clauses falsified by the solutions. Figure \ref{fig.2} shows the distribution of these solutions over the number of falsified clauses of each solver for the six representative instances, in which each point $(x, y)$ in a curve represents the fact that the PB function returned $y$ solutions falsifying $x$ clauses within 300 seconds for the corresponding solver.

Two observations can be made from Figure \ref{fig.2}, explaining the performance of IPBMR and IPBMR-noRandom over IPBR and IPBMR-noBreak.

\begin{itemize}
\item The solutions returned by the PB function falsify substantially fewer clauses in IPBMR and IPBMR-noRandom than in IPBR and IPBMR-noBreak, because the mutation and Path-Relinking strategies allow IPBMR and IPBMR-noRandom to focus only on high quality solutions.
\item With the same cutoff time of 300 seconds, IPBMR and IPBMR-noRandom call much more often the PB function than IPBR and IPBMR-noBreak, meaning that the PB function is executed much faster in IPBMR and IPBMR-noRandom, so that PBMR and IPBMR-noRandom can explore much more solutions returned by the PB function than IPBR and IPBMR-noBreak. This observation can be explained as follows.  On the one hand, the Path-Breaking strategy makes the PB function carries out less calculations in IPBMR and IPBMR-noRandom than in IPBMR-noBreak. On the other hand, although IPBR also uses the Path-Breaking strategy, it restarts the search process each time it encounters a local minimum solution, while IPBMR and IPBMR-noRandom apply the mutation strategies to the local minimum solution, allowing to keep a part of its good quality. In other words, while IPBR works on a randomly generated solution, IPBMR and IPBMR-noRandom work on a mutated local minimum solution. The difference is that the breaking condition in the PB function can be satisfied earlier when working on a mutated local minimum solution than on a randomly generated solution. Consequently, the calculations of the PB function are less time-consuming in IPBMR and IPBMR-noRandom than in IPBR.
\end{itemize}

\section{Conclusions and Future Work}

We proposed a new effective local search algorithm called IPBMR for the MAX-SAT, based on the classical Path-Relinking method. The performance of IPBMR comes from a careful combination of three components:

\begin{itemize}
\item A Path-Breaking strategy, which significantly increases the probability to find a good solution by avoiding exploring unpromising regions in the search space.
\item A new variable picking heuristic to generate paths between two elite solutions, allowing to avoid premature local minimum solutions.
\item A weak and a strong local optimum solution mutation strategies, allowing to keep parts of good properties of a local optimum solution, so that the search is diversified but is kept in the promising regions of the search space, which is better than a complete restart of the search.
\end{itemize}

We conducted an in-depth empirical investigation to identify and explain the effect of the three components, and to compare IPBMR with two state-of-the-art local search algorithms, CCLS and Swcca-ms, which do not use Path-Relinking, but focus on falsified clauses to pick the next variable to flip. Experimental results show that IPBMR significantly outperform CCLS and Swcca-ms, indicating that the Path-Relinking method reinforced with the three components is very effective for the MAX-SAT.

The mutation strategies of IPBMR, which aim at diversifying the search but keeping good properties of a local optimum solution, are based on flipping a randomly chosen subset of variables in the solution. In the future, we plan to use some machine learning approach to accurately identify the good properties in the local optimum solution, so that the mutation can better keep these good properties. We believe that this is a promising direction to improve the performance of IPBMR on industrial MAX-SAT instances.

\section{Acknowledgement}
This work was supported by National Natural Science Foundation of China (61472147, 61772219, 61602196).

\bibliographystyle{acm}
\bibliography{myreference}

\end{document}